# State Space Models for Extractive Summarization in Low Resource Scenarios

**Nisrine Ait Khayi**
University of Memphis
ntkhynyn@memphis.edu

## Abstract

Extractive summarization involves selecting the most relevant sentences from a text. Recently, researchers have focused on advancing methods to improve state-of-the-art results in low-resource settings. Motivated by these advancements, we propose the MPoincareSum method. This method applies the Mamba state space model to generate the semantics of reviews and sentences, which are then concatenated. A Poincare compression is used to select the most meaningful features, followed by the application of a linear layer to predict sentence relevance based on the corresponding review. Finally, we paraphrase the relevant sentences to create the final summary. To evaluate the effectiveness of MPoincareSum, we conducted extensive experiments using the Amazon review dataset. The performance of the method was assessed using ROUGE scores. The experimental results demonstrate that MPoincareSum outperforms several existing approaches in the literature.

## Introduction

Extractive text summarization is a fundamental Natural Language Processing (NLP) task that involves selecting the most salient sentences from a text while preserving its meaning. This process typically consists of two sequential steps (Zhou et al., 2018). The first step involves predicting the relevance of sentences using a deep learning model (Gu et al., 2022; Xie et al., 2022; Jia et al., 2020; Narayan et al., 2020). The second step ranks the sentences based on their relevance scores and selects the top-ranked ones.

However, state-of-the-art models for extractive summarization are often data-hungry, with their complexity increasing in proportion to the large number of parameters. This makes extractive summarization particularly challenging in low-resource scenarios, such as those involving limited annotated datasets or computational resources. These limitations can negatively impact the language representation of sentences and reduce model performance. To address these challenges, several techniques have been proposed, including data augmentation, distant supervision (Liang et al., 2024), embeddings

| **Gold** | fairly strong chemical taste |
|---|---|
| **Ours** | my first cup began with a fairly pronounced chemical taste. As I proceeded through the cup, the effect went away somewhat, but completely. |
| **Reviews** | my first cup began with a fairly pronounced chemical taste as I proceeded through the cup the effect went away somewhat but entirely; I don't normally drink flavored coffee, but this is not something I would choose to drink again or feel comfortable serving to friends. |

Table1. Example of summaries generated by our model and the golden summaries in the Amazon review dataset.

using pre-trained language models (Bajaj et al., 2021), domain adaptation of language models (Yu et al., 2021), and Meta-Learning (Chen et al., 2021). Biljon et al. (2020) demonstrated that low-medium pre-trained language models with fewer parameters perform better on few-shot tasks.

One issue with transformer models is their inefficiency when dealing with long sequences. To alleviate this, Gu et al. (2024) proposed a State Space Model (SSM) with linear scaling in sequence length, which has shown improved performance in zero-shot tasks compared to attention-based models. Akbik et al. (2018, 2019b) demonstrated that powerful representations can be generated by concatenating high-resource embeddings from a general domain with low-resource embeddings from a target domain. Lange et al. (2020b) proved that domain-invariant representations can be generated by training embeddings on diverse domains using an adversarial discriminator to distinguish between embedding spaces. Meta-learning algorithms (Dou et al., 2019) aim to learn effective initializations for fine-tuning across various tasks with minimal training data.

In this paper, we frame the task of extractive summarization for the Amazon Review dataset (He and McAuley, 2016) in a low-resource setting as a binary classification of sentences and reviews, followed by paraphrasing the extractive summary to align with human summarization. Motivated by prior successes, our contributions are as follows:

2  The semantics of sentences are generated using the State Space Model Mamba (Gu et al., 2024), which employs a selective scan algorithm to compress information, a HiPPPO initialization matrix to capture long-range dependencies, and a hardware-aware algorithm to accelerate computations.
3  The generated semantics are further compressed using spectral clustering (Macgregor et al., 2023) and Poincaré distance (Klein & Hilbert, 1999). The distances between embeddings and centroids are used as features.
4  We explore parameter-efficient fine-tuning (PEFT) through LoRA (Low-Rank Adaptation) (Hu et al., 2022), which integrates low-rank decomposition matrices into the model layers.

## 5  Related Work

State-of-the-art models in the extractive summarization research area heavily rely on the quality of sentence semantics. However, limited training datasets or computational resources can lead to a decline in performance. Recently, low-resource extractive summarization has gained increased attention. Yuan and colleagues (2023) applied disentangled representation learning to extractive summarization, separating context and pattern information (e.g., sentence position, n-gram tokens) to enhance generalization in low-resource settings. They proposed two loss functions for encoding and disentangling sentence representations into context and pattern components: 1) an adversarial objective, and 2) a mutual information minimization objective.

To reduce the need for large amounts of training data, Tang and colleagues (2023) introduced ParaSum, which reformulates extractive summarization as paraphrasing. This aligns the task with the self-supervised next-sentence prediction objective of pre-trained language models. Experimental results demonstrated the effectiveness of this approach in low-resource settings. Brazinskas and colleagues (2020) proposed a few-shot framework for abstractive opinion summarization, enabling successful cross-domain adaptation. In the first stage, a conditional transformer is trained to generate product reviews conditioned on summary-related review properties. In the second stage, they fine-tune a module to predict these summary properties. This framework outperforms many extractive and abstractive summarization models.

In this work, we build upon these advancements in enhancing sentence semantics to improve extractive summarization performance in low-resource settings. We propose using the State Space Mamba model, combined with Poincaré compression and Spectral clustering, to generate sentence features. ur approach is evaluated using the Amazon review dataset.

## 6  Method

In this section, we describe the key components of our proposed model for classifying sentences as relevant or irrelevant for the final summary. First, we outline the main modules of the Mamba block: 1) Recurrent State Space Model (SSM) through discretization, 2) HiPPO initialization on matrix A to capture long-range dependencies, 3) Selective scan algorithm for compressing information, and 4) Hardware-aware algorithm to accelerate computations. Next, we explain the Poincaré compression method used to extract the most meaningful features from the Mamba-based embeddings. Finally, we provide details on the Low-Rank Adaptation (LoRA) technique, which incorporates low-rank decomposition matrices into the Mamba model.

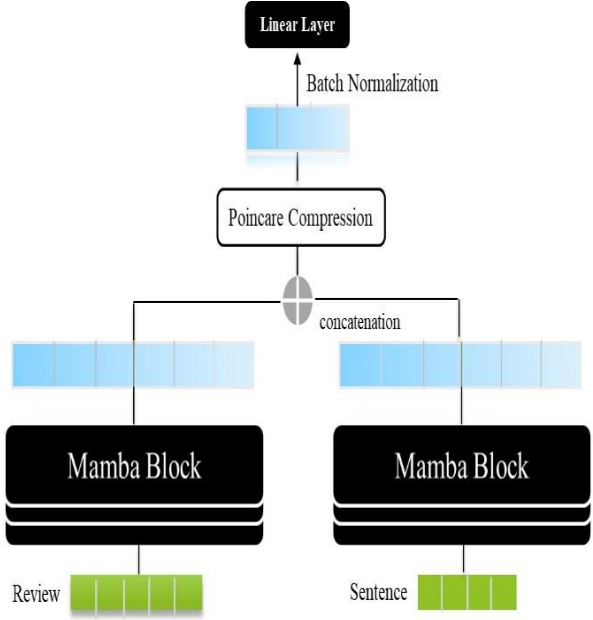

Figure 1. Overview of the proposed model to classify sentences as being relevant or not for the summary. The components of the models are: 1- Mamba Block, 2-Poincare compression, 3- Batch Normalization, 4- Linear Layer

### 3.1 Task Formulation

We formulate the extractive summarization as a binary classification of a sentence of being relevant or not, given its review. Then, we paraphrase the relevant sentences using BART, sequence-to-sequence model (Lewis et al., 2020).

- Given a review of N sentences $[s_1, s_2..., s_N]$, we pass it, with each sentence, to the State Space Mamba model. Then, we collect the mean of the last hidden state as the representation of the review $H_r$ and the representation of the sentence $H_s$. Then, we concatenate them as $H_{rs}$, and we apply Spectral clustering using Poincare distance to generate the review and the sentence features $F_{rs}$.

$$F_{rs_i} = d_{poincare}(H_{rs_i}, C_j) \qquad (1)$$

where $d_{poincare}$ is the Poincare distance between the i[th] vector of the $H_{rs}$ embedding and the j[th] centroid.

- Then, we apply a batch normalization and linear layer to predict the class label y for the sentence $s_i$:

$$y = w.H + b \qquad (2)$$

where w is the weight vector, b is the bias and y is the final output.

- After extracting the relevant sentences $[s_1, s_2 ..., s_m]$, we paraphrase them using BART.

$$sum = BART\ (s_1) \oplus ... \oplus BART\ (s_m) \qquad (3)$$

### 3.2 Selective Space Model

The most recent advances in extractive summarization are based on Transformer-based models. However, a major flaw of Transformers is the slow inference due to the self-attention computations when the sequence length of the input increases. To overcome this drawback, Mamba SSM (Gu et al., 2023) architecture parallelizes training as Transformers and performs inference that scales linearly with the sequence length. This can be an effective model for low-resource scenarios. Mamba SSM has the following properties:

**Recurrent SSM**

A recurrent technique is used to compute a discretized version of the Mamba model. At each time step, the model calculates how the current input $(Bx_k)$ influences the previous state $(Ah_{k-1})$ and then calculates the predicted output $(Ch_k)$.

$$\bar{A} = exp(\Delta A) \qquad (4)$$

$$\bar{B} = (\Delta A)^{-1}(exp(\Delta A) - I) \cdot \Delta B \qquad (5)$$

$$h_k = \bar{A}h_{k-1} + \bar{B}x_k \qquad (6)$$

$$y_k = Ch_k \qquad (7)$$

A, B, and C are the parameters of the model.

This technique has the advantage of fast inference and the disadvantage of slow training.

**HiPPO initialization on matrix A**

The idea behind HiPPO (High order Polynomial Projection Operators) (Gu et

al.,2020) is to produce a hidden state that memorizes history. This is computed by tracking the coefficients of a Legendre polynomial that approximates all the history. HiPPO is applied to the recurrent presentation to handle long-range dependencies.

**Selective scan algorithm**

Mamba compresses meaningful data into the state by incorporating the sequence length and batch size into matrices B and C.

**Hardware aware algorithm**

Mamba reduces the number of times to copy information between SRAM (Static Random Access Memory) and DRAM (Dynamic Random Access Memory). It does it through kernel fusion that continuously performs computation until it's done. The algorithm also recomputes the intermediate states during the backward pass which is less costly.

### 3.3 Poincare Compression

The semantic vectors used by the Mamba model are dense and high-dimensional, which can slow training and reduce the model's generalizability in low-resource settings. To address this challenge, various compression techniques have been proposed. In this work, we aim to reduce the dimensionality of the Mamba-based embeddings and select the most meaningful vectors using Spectral clustering and Poincaré distance, as outlined in the following algorithm.

---
**Algorithm Poincare Compression**

**Input:** embeddings $\{e_i\}_{i=1}^{m}$ and dimensionality d centroids $\{c_j\}_{j=1}^{n}$

for $i = 1,2,\ldots,m$ do
  for j= $1,2,\ldots,n$ do
    $h_{i,j} = d_{Poincare}(e_i, c_j)$
  end for
end for

**Output:** Poincare-Spectral embeddings $\{h\}_{i=m, j=n}$

---

Given Mamba-based embeddings, we apply Spectral clustering. Then, we compute the Poincare distance between each embedding vector and the centroids to generate the features. Spectral clustering and Poincare distance are described in the next sections.

**Spectral Clustering**

Spectral clustering (Ng et al.,2001) is robust for high-dimensional data as it uses the distance on a graph, which is more meaningful compared to Euclidean distance. Moreover, it is simple, fast, and effective.

Given a graph $G$ with $n$ vertices and $k$ clusters, the spectral clustering algorithm consists of these two steps:

1. Embed the vertices of $G$ into $\mathbb{R}^k$ according to $k$ eigenvectors of the graph Laplacian matrix
2. Apply the k-means clustering algorithm to partition the vertices into $k$ clusters.

**Poincare Distance**

Nickel et al (2017) have demonstrated that the embeddings in Poincare space outperform the embeddings in Euclidean space as they account for the hierarchal structure of the text. Motivated by this success, we embedded the Mamba-based semantics in the Poincare space.

The Poincare distance $d_{Poincare}(a, b)$ between two points a and b is given by this formula:

$$d_{Poincare}(a, b) = 1 + 2|a - b|^2(1 - |a|^2)(1 - |b|^2) \quad (8)$$

### 3.4 LoRA (Low-rank adaptation)

Many research findings (Jukic et al.,2023) affirm the superiority of Parameter Efficient Fine Tuning (PEFT) over Full-Fine Tuning (FFT) in low-resource settings. Inspired by this, we apply LoRA (Hu et al., 2021) to our proposed model to improve the performance of our extractive summarization in a low-resource setting. The LoRA technique freezes the pre-trained model weights and incorporates trainable rank decomposition matrices into the model's layers. This results in reducing the trainable parameters for the downstream tasks.

### 3.5 Paraphrasing

After selecting the relevant sentences in the review, we paraphrase them using the BART model (Lewis et al., 2020). It is a denoising autoencoder for pretraining sequence-to-sequence models. BART is trained by:

1. Corrupting text with an arbitrary noising function
2. Learning a model to reconstruct the original text.

The used transformations are: (1) token masking, (2) token deletion, (3) text infilling, (4) sentence permutation, and (5) document rotation.

# 7 Experiments and Results

To evaluate the strength of our proposed method, we have conducted several experiments using the Amazon review dataset.

## 4.1 Amazon Review Dataset

We used the Amazon customer review dataset (He and McAuley, 2016), which includes text reviews, summaries, user information, and product details. For our study, we sampled 136 product reviews for training and 73 for testing. The text reviews were split into individual sentences, which were then annotated as relevant or irrelevant. The annotation process was based on ROUGE scores and the semantic similarity between each sentence and its corresponding review..

## 4.2 Experiments Settings

We performed our experiments using a P5000 GPU and an A4000 GPU, both equipped with 30 GB of RAM. The proposed model was implemented using HuggingFace's library, with the "mamba-130m-hf" version of the Mamba model and tokenizer. The maximum sequence length for the Mamba model was set to 128. We used the AdamW optimizer (Loshchilov et al., 2017) with a learning rate of 2e-5 and a weight decay of 0.5. A one-cycle learning rate scheduler (Smith et al., 2015) was employed to adjust the learning rate during training. To address overfitting in our low-resource setting, we applied a dropout rate of 0.5 and used BatchNorm1d for regularization. For LoRA configuration, we set lora_alpha to 32 and lora_dropout to 0.1.

The dataset was preprocessed by converting text to lowercase, removing URLs, digits, and punctuation, handling missing data via imputation, and normalizing the data before applying Spectral clustering. We evaluated our model using ROUGE-1, ROUGE-2, and ROUGE-L metrics. Each experiment was repeated multiple times, and we selected the best-performing model.

## 4.3 Summarization of Results and Analysis

Table 3 presents the empirical results of our proposed method on the Amazon review dataset, comparing it with several existing text summarization models from the literature. We report the ROUGE-1 (R1), ROUGE-2 (R2), and ROUGE-L (RL) scores. As shown in the table, our method, MPoincareSum, outperforms several models, including Copycat (Brazinskas et al., 2020), MeanSum (Chu and Liu, 2019), and LexRank (Erkan and Radev, 2004), in terms of R2 and RL scores. This highlights the effectiveness of State Space models, Poincaré compression, and Parameter-Efficient Fine-Tuning for extractive summarization in low-resource settings.

However, FewSum (Bražinskas et al., 2020) outperforms MPoincareSum in terms of R1 by 0.158. This can be attributed to their use of fluent and informative review summaries generated by a conditional transformer language model. In contrast, Amazon reviews are often unstructured and contain informal language, which may explain the performance gap.

|  | R1 | R2 | RL |
|---|---|---|---|
| FewSum | **0.372** | 0.099 | **0.227** |
| Copycat | 0.281 | 0.058 | 0.183 |
| MeanSum | 0.275 | 0.035 | 0.160 |
| LexRank | 0.269 | 0.049 | 0.161 |
| **MPoincareSum** | 0.214 | **0.12** | **0.2** |

Table 3 Experiments Results

# 8 Ablation Study

We conducted an ablation study to evaluate the impact of different components in our model. First, we experimented without the Poincaré compression module, meaning the semantics of sentences and reviews were encoded using the SSM Mamba model without dimensionality reduction. The results in Table 4 show that removing the Poincaré compression module leads to a significant drop in performance, indicating that compressing the SSM Mamba embeddings enhances the extractive summarization performance in low resource settings.

Next, we experimented without the sequence routing of the Mamba model, replacing it with a transformer model, such as BERT (Devlin et al., 2019), for encoding the semantics of sentences and reviews. The results in Table 4 reveal that the self-attention mechanism of a transformer and the sequence routing of a state space model have similar impacts on extractive summarization in low-resource settings.

Finally, we explored the effect of Parameter-Efficient Fine-Tuning (PEFT) using LoRA by comparing it to full fine-tuning (FFT) of our proposed model. As shown in Table 4, the ROUGE scores significantly dropped when using full fine-tuning, suggesting that PEFT LoRA improves the performance of of extractive summarization in low-resource settings.

## Conclusion

We presented the MPoincareSum method for extractive summarization in low-resource settings, applied to the Amazon review dataset. Our proposed approach involves extracting semantic features from reviews and sentences using a State Space Mamba model. These features are then concatenated, and we apply Spectral clustering to select the most meaningful ones. Next, we compute the Poincaré distance between the embedding vectors and the centroids. A linear layer is then used to predict the relevance of each sentence. To reduce the model's parameters, we incorporate the LoRA technique. Finally, we paraphrase the relevant sentences using the BART sequence-to-sequence model to include them in the final summary.

To evaluate the performance of MPoincareSum, we conducted several experiments. The empirical results demonstrate the effectiveness of State Space models and Poincaré compression for extractive summarization in low-resource settings, yielding competitive ROUGE scores.

Looking ahead, we plan to further improve ROUGE scores in low-resource extractive summarization by experimenting with Mixture of Experts (MoE) models, which offer enhanced training efficiency without compromising inference performance

## Ethics Statement

Our research work uses Amazon reviews, a publicly available dataset. Our annotation process does not harm the intellectual property of the original authors of the dataset. The scientific artifacts used are available for research with licenses, including ROUGE, Hugging Face, and PEFT. Their use does not violate their intended use. Our task is a well-defined NLP task which is text summarization. Thus, there are no potential risks of this work.